\documentclass[conference]{IEEEtran}
\IEEEoverridecommandlockouts
\usepackage{cite}
\usepackage{amsmath,amssymb,amsfonts}
\usepackage{algorithmic}
\usepackage{graphicx}
\usepackage{subcaption}

\usepackage{tikz}
\usetikzlibrary{positioning}
\usepackage{tikz}
\usepackage{tikzscale}
\usepackage{xspace}
\usepackage{pgfplots}
\usepgfplotslibrary{groupplots}
\usepackage{multirow}

\usepackage{textcomp}
\usepackage{xcolor}

\usepackage{url}
\usepackage{multicol}
\usepackage{multirow}
\usepackage{float}
\usepackage{hyperref}  

\usepackage{graphicx,xfp}
\usepackage{multicol}
\usepackage{multirow}
\usepackage{color}
\usepackage{xcolor}
\usepackage{booktabs}
\usepackage{graphicx}
\usepackage{nicefrac}
\usepackage{amsmath}

\usepackage{array}
\newcolumntype{P}[1]{>{\centering\arraybackslash}p{#1}}

\usepackage{graphicx}
\usepackage{fancyhdr}
\usepackage{amssymb}
\usepackage{url}
\usepackage{color}
\usepackage{tikz}
\usetikzlibrary{positioning}
\usepackage{pgfplots}
\usepackage{marvosym}

\usetikzlibrary{external}
\def\BibTeX{{\rm B\kern-.05em{\sc i\kern-.025em b}\kern-.08em
    T\kern-.1667em\lower.7ex\hbox{E}\kern-.125emX}}
\begin{document}

\title{Informative Initialization and Kernel Selection Improves t-SNE
  for Biological Sequences}


\author{\IEEEauthorblockN{Prakash Chourasia, Sarwan Ali, Murray Patterson*}
\IEEEauthorblockA{
\textit{Department of Computer Science, Georgia State University}\\
Atlanta, GA, USA \\
\{pchourasia1,sali85\}@student.gsu.edu, mpatterson30@gsu.edu} \\
\thanks{*Corresponding Author}
\thanks{2022 IEEE International Conference on Big Data (Big Data) | 978-1-6654-8045-1/22/\$31.00 \copyright 2022 European Union}
}


\maketitle

\begin{abstract}

The t-distributed stochastic neighbor embedding (t-SNE) is a method
for interpreting high dimensional (HD) data by mapping each point to a
low dimensional (LD) space (usually two-dimensional).
It seeks to retain the structure of the data.  An important component
of the t-SNE algorithm is the initialization procedure, which begins
with the random initialization of an LD vector. Points in this initial
vector are then updated to minimize the loss function (the KL
divergence) iteratively using gradient descent. This leads comparable
points to attract one another while pushing dissimilar points apart.
We believe that, by default, these algorithms
should employ some form of informative initialization. Another
essential component of the t-SNE is using a kernel matrix, a
similarity matrix comprising the pairwise distances among the
sequences. For t-SNE-based visualization, the Gaussian kernel is
employed by default in the literature. However, we show that kernel
selection can also play a crucial role in the performance of t-SNE.

In this work, we assess the performance of t-SNE with various
alternative initialization methods and kernels, using four different
sets, out of which three are biological sequences (nucleotide, protein, etc.) datasets obtained
from various sources, such as the well-known GISAID database for sequences of the SARS-CoV-2 virus. 
We perform subjective and objective
assessments of these alternatives. We use the resulting t-SNE plots
and $k$-ary neighborhood agreement ($k$-ANA) to evaluate and compare the proposed methods with
the baselines. We show that by using different techniques, such as
informed initialization and kernel matrix selection, that t-SNE
performs significantly better. 
Moreover, we show that t-SNE also takes fewer iterations to converge faster with more intelligent initialization.

\end{abstract}

\begin{IEEEkeywords}
t-SNE, Visualization, Initialization, Kernel Matrix, Biological
Sequences
\end{IEEEkeywords}

\maketitle

\section{Introduction}
The quantity of high dimensional data sets in genomic sequencing
necessitates dimensionality reduction techniques and ways to aid in
creating data visualizations. Principal component analysis (PCA) and
independent component analysis (ICA) are standard
dimensionality reduction techniques. Among them, the t-distributed
stochastic neighbor embedding (t-SNE) introduced by van der Maaten and
Hinton (2008)~\cite{hinton2008visualizing} has recently gained
popularity, especially in the natural sciences, due to its
ability to handle large amounts of data and its use for dimensionality
reduction while preserving the structure of data. It is typically
used to create a two-dimensional embedding of high-dimensional data to
simplify viewing while keeping its overall structure. Despite its
enormous empirical success, the theory behind t-SNE remains to be seen.

This paper demonstrates the effect of informed initialization on the
step-wise process (incremental performance, convergence, etc.) of the
t-SNE algorithm.
We also review the impact of different kernels on t-SNE for various
types of biological (nucleotide, protein, etc.) sequence data,
including SARS-CoV-2\footnote{The SARS-CoV-2 virus is the cause for
  the global COVID-19 pandemic.}  sequences. The vast global spread of
pandemics like COVID-19 provided the impetus for this research,
pushing viral sequence analysis into the ``Big Data'' realm.
This leads to challenges in reducing high-dimensional data to
low-dimensional space, not just to conserve computational resources
when using it in cutting-edge methods like machine learning, but also
to improve visualization.

The severe acute respiratory syndrome coronavirus 2 (SARS-CoV-2) virus
is a member of the genus Betacoronavirus, and its genetic material is
a single positive-strand RNA~\cite{bai2021overview}. Its viral genome
(Gene ID—MN908947) has roughly $29,881$
nucleotides~\cite{raman2021covid}. The virus has a double-layered
lipid envelope with four structural proteins, S, M, E, and N, in its
structure.
Most of the mutations related to SARS-CoV-2 occur in the spike region~\cite{walls2020structure}.

In this work, we use four datasets, of which three sets are biological
sequences, to assess the effect of informed initialization and kernel
selection on t-SNE. Firstly, we use a toy dataset as a proof of
concept. Among others, there is a set of 7000 SARS-CoV-2 spike protein
sequences
obtained from the well-known
GISAID\footnote{\url{https://gisaid.org/}} database.  The COVID-19
pandemic has generated a renewed interest in the larger family
Coronaviridae, among which SARS-CoV-2 is a member, along with others
such as the original severe acute respiratory syndrome and the
middle-eastern respiratory syndrome
coronaviruses~\cite{ali2022pwm2vec}. Since the family Coronaviridae
affects a wider variety of hosts, it is often of interest to
characterize such viral sequences with the primary host that it
affects. Hence, the third dataset used contains the host for each
sequence. Finally, sometimes researchers only obtain such data in the
form of short reads (when no reference genome is available, for
example). To simulate this scenario, our fourth dataset is obtained by
taking a set of $\approx$10K full-length nucleotide sequences of
SARS-CoV-2, and simulating the process of sequencing, Illumina reads
from each such nucleotide sequence.


In this paper, our contributions are the following:
\begin{enumerate}
\item We convert the biological sequences to a fixed-length
  numerical representation, afterward, we compute different kernel matrices and
  show their effect along with several initialization methods on the
  quality of t-SNE.
\item We show that the Kernel selection can play an important role and should be considered carefully rather than using the typical Gaussian kernel for t-SNE computation. 
\item We show that random initialization is inefficient for t-SNE computation. Alternatively, the Ensemble approach is
  a better choice to start with as an initial solution than the
  Random, Principal Component Analysis (PCA), and Independent
  Component Analysis (ICA) approaches.
\item We evaluate the performance of the t-SNE using subjective (t-SNE
  plots) and objective ($AUC_{RNX}$
  )
  criteria, and report results for different kernel computation
  methods along with different initialization approaches.
\item We show that our proposed setting that includes the use of
  Laplacian kernel, along with ensemble initialization, outperforms the
  typical Gaussian and Isolation kernel-based methods in terms of
  $AUC_{RNX}$.
\end{enumerate}

The rest of the paper is organized as follows:
Section~\ref{sec_related_work} contains the related work for the t-SNE
computation problem. Our proposed solution is given in
Section~\ref{sec_tsne_compute}. The dataset statistics, along with the
experimental setup details, are given in
Section~\ref{sec_experimental_setup}. Our results are reported in
Section~\ref{sec_results_discussion}. Finally, we conclude the paper
in Section~\ref{sec_conclusion}.

\section{Related Work}
\label{sec_related_work}

The task of data visualization is crucial. This work has been made
easier using t-SNE, which was first introduced
in~\cite{van2008visualizing}. Authors in~\cite{ali2021k} use it to
show distinct variations in coronavirus protein sequence data. It was
also discovered that employing k-means to cluster SARS-CoV-2 protein
sequences is similar to the patterns shown in t-SNE
plots~\cite{ali2021effective,tayebi2021robust}. Authors
in~\cite{shaham2017stochastic} present a theoretical feature of SNE (a
forerunner to t-SNE) requiring that global minimizers quantitatively
separate clusters. However, their finding is only nontrivial when the
number of clusters is much greater than the number of points per
cluster, which is not a reasonable assumption in general.

Authors in~\cite{kobak2021initialization} show the importance of
initialization in UMAP and t-SNE. With informed initialization, t-SNE
performs as well as UMAP. However, they have yet to consider the
importance of kernel selection. In~\cite{saha2017see}, a decentralized
data stochastic neighbor embedding (dSNE) is developed, which is
beneficial for visualizing decentralized
data. In~\cite{saha2021privacy}, authors propose a deferentially
private dSNE (DP-dSNE) variant. Both dSNE and DP-dSNE use
similarity to map the data points from distinct places. The dSNE and
DP-dSNE approaches come in handy when data cannot be shared due to
privacy issues. Although current t-SNE approaches are effective on
popular datasets like MNIST, it is unclear if they can be applied to
large datasets of biological sequences.

\section{Proposed Approach}\label{sec_tsne_compute}

In this section, we first discuss the method to convert the sequences into a fixed-length numerical representation. We then describe the different kernel computation methods. Finally, we describe initialization approaches for the t-SNE.

\subsection{Numerical Embedding Generation}
Since t-SNE operates on the numerical vectors/embeddings, the first step is to convert the sequences into fixed-length numerical representations. For this purpose, we use a recently proposed method called Spike2Vec~\cite{ali2021spike2vec}.
Given a sequence, Spike2Vec generates a fixed-length numerical representation using the concept of $k$-mers (also called n-gram).
It uses the idea of the sliding window to generate substrings (called mers) of length $k$ (size of the window).
For a set of $k$-mer for a biological sequence, we generate a feature vector of length $\vert \Sigma \vert ^k$ (where $\Sigma$ corresponds to the set of alphabets ``amino acids" or nucleotide in the sequence), which contains the frequency/count of each $k$-mer.

\subsection{t-SNE computation pipeline}
\begin{figure}[h!]
  \centering
  \includegraphics[scale=0.15]{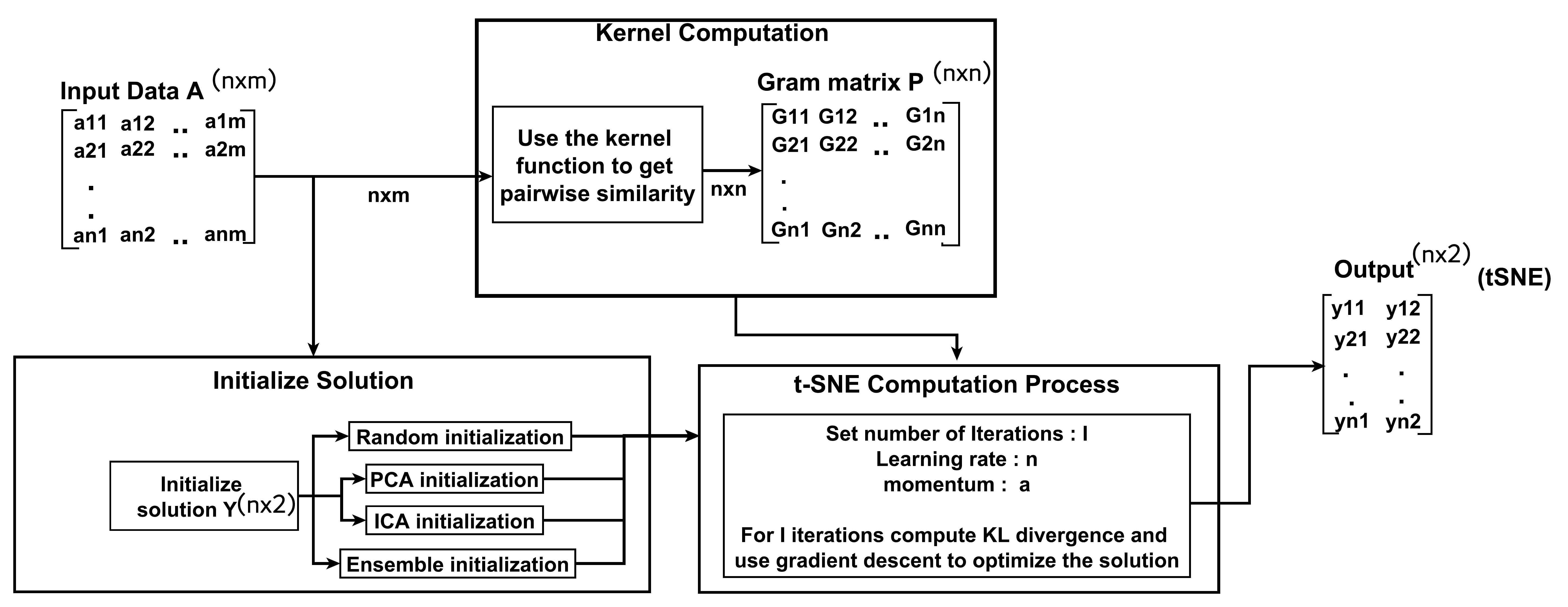}
  \caption{Overall pipeline of our model. Given the input data (A), we
    first compute the kernel matrix (P) using different kernel
    functions. We then Initialize a solution (Y) using different initialization approaches. Later we apply Gradient
    Descent to get the low dimensional representation of the data.}
  \label{process_flow_chart}
\end{figure}

The overall pipeline of our model is shown in Figure~\ref{process_flow_chart}. In the t-SNE computation pipeline, there are three steps involved, namely (1) Kernel matrix computation, (2) Solution Initialization, and (3) Gradient Descent.  

\subsubsection{Kernel Functions}
For kernel matrix computation, we use four different kernel functions, which are described below, and also the process is shown in the first box of Figure~\ref{process_flow_chart}.

\paragraph{Gaussian Kernel~\cite{hinton2002stochastic}}The N-dimensional Gaussian kernel is computed for any number of points by computing each Kernel coordinates and their distance to the chosen points, then taking the Gaussian function of these distances. In 2D, the Gaussian Kernel is defined as:
$
\mathcal{K}\left(x, y\right)=\exp \left(\frac{-\left\|x-y\right\|^{2}}{2 \sigma^{2}}\right)
$, 
where the width of the Kernel is defined by $sigma$, we computed the difference using Squared Euclidean distance with a perplexity value of $250$ for our trials, and we are tweaking $sigma$. The Gaussian Kernel is a default Kernel used in the t-SNE implementation. We use this as a baseline to evaluate the performance of other Kernel.

\paragraph{Isolation Kernel ~\cite{zhu2021improving}}It is a data-dependent kernel that is quick to compute because it just has one parameter. It adjusts to the local density distribution, unlike the Gaussian kernel. For given two points $x$, $y \in \mathrm{R}^d$, the Isolation kernel of $x$ and $y$ with respect to 
$\mathrm{D}$ is defined to be the expectation taken over the probability distribution on all partitioning $\mathrm{H} \in \mathrm{H_\psi} (\mathrm{D})$ which both $x$ and $y$ fall into the same isolating partition $\theta[z]\in \mathrm{H}, z\in D$ given by:
$
K_{\psi}(x, y \mid D)=\mathbb{E}_{\mathbb{H}_{\psi}(D)}[\mathbb{1}(x, y \in \theta[z] \mid \theta[z] \in H)]
$.

\paragraph{Laplacian Kernel~\cite{hajiaboli2011edge}} The Laplacian kernel function gives the relationship between two input feature vectors $x$ and $y$ in infinite-dimension space. Its value ranges between $0$ and $1$ where, $1$ denotes $x$ and $y$ are similar. Can be denoted by the following:   
$
k(x, y)=\exp \left(-\frac{1}{2 \sigma^2}\|x-y\|_{1}\right)
$, where $\mathrm{x}$ and $\mathrm{y}$ are the input vectors and $\|x-y\|_{1}$ is the Manhattan distance between the input vectors.

\paragraph{Approximate Kernel~\cite{ali2022efficient}}
This kernel offers a mechanism to compare two sequences' similarities by computing a spectrum based on the number of matches $k$ and mismatches $m$ between $k$-mers of two sequences. Given a pair of the sequence $X$ and $Y$, the spectrum (frequency count-based vector of $k$-mers) between any two sequences will have a length equal to the size of  more significant length sequences between $X$ and $Y$, which will contain the counts of matches and mismatches between characters in the sequence $k$-mers. 
This method computes $n \times n$ kernel matrix based on the dot product of vectors. We use $k=3$ and $m=0$ for computing this kernel.

\subsubsection{Solution Initialization}
It is an important component of the t-SNE computation process. For the initialization of t-SNE, we use the following four techniques, also shown in the second box of Figure~\ref{process_flow_chart}. 

\paragraph{Random Initialization}
Random numbers are assigned to the $N \times 2$ matrix, where N is the total number of sequences.

\paragraph{Principle Component Analyses (PCA)~\cite{wold1987principal1}}
Singular Value Decomposition reduces the data's dimensionality and projects it to a lower dimensional environment. We are not using PCA as a dimensionality reduction method here. Rather, we use it to get intelligent initial t-SNE vectors(initialization).
Here we get $N \times 2$ matrix as output when numerical embedding is provided as input to PCA. This low-dimensional output is used to initialize the solution for the t-SNE algorithm.

\paragraph{Independent Component Analyses (ICA)~\cite{hyvarinen2000independent}}
ICA establishes two essential assumptions, i.e., statistical independence and non-Gaussian distribution property, among the components. In a semantic sense, the information about $x$ (original data) does not provide information about $y$ (new data comprised of independent components), and vice versa. 
Its goal is to separate the data by transforming the input space into a maximally independent basis.
More formally, this corresponds to $ p(x, y)=p(x) p(y) $, where the probability distribution of x is represented by p(x), and p(x,y) represents the joint distribution of x and y.

\paragraph{Ensemble}
We took the average for $N \times 2$ matrices generated by PCA and ICA in this method. Initialize the solution in t-SNE with the generated averaged matrix.

\subsubsection{Gradient Descent}
Ultimately, we use gradient descent to optimize the t-SNE-based 2D representation. As shown in the last part of Figure~\ref{process_flow_chart}, we set the parameters like learning rate, momentum, and Iteration. KL divergence is used to measure the distance between two distributions (Distribution among distances in the data points in LD and HD). Taking the cost function's derivative and calculating gradient descent, we keep updating the initial solution $Y$ to get the optimal solution. Finally, we apply t-distribution on low-dimension distribution, which gives us a longer tail to give better visualization.


\section{Experimental Setup}
\label{sec_experimental_setup}
In this section, we first discuss the dataset statistics. After that, we report the goodness metrics used to evaluate the performance of t-SNE. All experiments are performed on Intel (R) Core i5 system with a 2.40 GHz processor and $32$ GB memory. 
The code and dataset used in this study are available online~\url{https://github.com/pchourasia1/tSNE_Informed_Initialization}.

\subsection{Dataset Statistics}\label{data_stats}
This section discusses the detailed description and statistics of datasets used for experiments.

\subsubsection{Circle Dataset}
Using a straightforward toy dataset, we can demonstrate the significance of random and non-random (informed) initialization. To create Kernel matrices, we collected $7000$ points from a circle with some additional Gaussian noise. 
\subsubsection{Spike Sequence Dataset}
The spike sequences are taken from GISAID~\footnote{\url{https://www.gisaid.org/}}, a well-known database. The entire retrieved sequences are $7000$, including $22$ lineages of virus. 
The detail of each lineage, i.e., name (count/distribution), in the dataset, is as follow B.1.1.7 (3369), B.1.617.2 (875), AY.4 (593), B.1.2 (333), B.1 (292), B.1.177 (243), P.1 (194), B.1.1 (163), B.1.429 (107), B.1.526 (104), AY.12 (101), B.1.160 (92), B.1.351 (81), B.1.427 (65), B.1.1.214 (64), B.1.1.519 (56), D.2 (55), B.1.221 (52), B.1.177.21 (47), B.1.258 (46), B.1.243 (36), and R.1 (32).

\subsubsection{Host Dataset}
This data is taken from~\cite{ali2022pwm2vec}, which comprised $5558$ spike sequences and coronavirus hosts as class labels. 
The count of each host (class label) is following: Bats (153), Bovines (88), Cats (123), Cattle (1), Equine (5), Fish (2), Humans (1813), Pangolins (21), Rats (26), Turtle (1), Weasel (994), Birds (374), Camels (297), Canis (40), Dolphins (7), Environment (1034), Hedgehog (15), Monkey (2), Python (2), Swines (558), and Unknown (2).
Length of Spike2Vec based feature vector is $13824$.

\subsubsection{Short Read Dataset}
This dataset is generated by taking $10181$ SARS-CoV-2 nucleotide
sequences and simulating short reads from each sequence using
inSilicoSeq~\footnote{\url{https://github.com/HadrienG/InSilicoSeq}} with the
miseq model (all other settings were default settings). 
The lineages/class labels (count/distribution) in this dataset is as follows: B.1.1.7 (2587), B.1.617.2 (1198), AY.4 (1167), AY.43 (412), AY.25 (275), B.1.2 (253), AY.44 (249), B.1 (200), B.1.177 (184), AY.3 (166), P.1 (137), B.1.1 (128), B.1.526 (112), AY.9 (108), AY.5 (96), AY.29 (86), AY.39 (85), B.1.429 (81), B.1.160 (79), Others (2578). There are $496$ unique lineages for $10181$ sequences.

\subsection{Evaluation Metrics}
We evaluate the t-SNE in two ways including subjective and objective evaluation. 
We plot t-SNE-based 2D visual representation for subjective evaluation to evaluate the performance. 
For objective evaluation, we use a technique called $k$-ary neighborhood agreement ($k$-ANA) method ~\cite{zhu2021improving} to analyze the t-SNE model's performance objectively. Given the original high dimensional (HD) data and the equivalent low dimensional (LD) data for this assessment method ($2$D data computed using the t-SNE approach), the $k$-ANA approach verifies the neighborhood agreement ($Q$) between HD and LD. It gets the intersection on the number of neighbors (for various $k$ nearest neighbors). In a formal setting:

\begin{equation}
\label{eqn_neignbor_agreement}
Q(k)=\sum_{i=1}^{n} \frac{1}{n k}\left \vert k N N\left(x_{i}\right) \cap k N N\left(x_{i}^{\prime}\right)\right \vert
\end{equation}

where $k$NN($x$) is the set of $k$ nearest neighbors of $x$ in high-dimensional and $k$NN($x'$) is the set of $k$ nearest neighbors of $x$ in low-dimensional (LD).
We employed a quality evaluation technique that quantifies neighborhood preservation and is denoted by R($k$), which evaluates on a scalar metric if neighbors are preserved~\cite{lee2015multi} in low dimensions using Equation~\ref{eqn_neignbor_agreement}. In a formal setting: 
$R(k)=\frac{(n-1) Q(k)-k}{n-1-k}$.
The value of R($k$) ranges from $0$ to $1$, with a higher score
indicating better neighborhood preservation in LD space. We computed
$R(k)$ for $k \in \{1,2,3,...,99\}$ in our experiment, then looked at
the area under the curve (AUC) created by $k$ and R($k$). Finally, we
compute the area under the $R(k)$ curve in the log plot
($AUC_{RNX}$)~\cite{lee2013type} to aggregate the performance of
various k-ANN. More formally,
$A U C_{RNX}=\frac{\Sigma_{k} \frac{R(k)}{k}}{\Sigma_{k} \frac{1}{k}}$,
where $AUC_{RNX}$ is an average quality weight for $k$ closest neighbors.

\section{Results and Discussion}\label{sec_results_discussion}
\begin{figure}[h!]
  \centering
  \begin{subfigure}{.24\textwidth}
  \centering
  \includegraphics[scale=0.12]{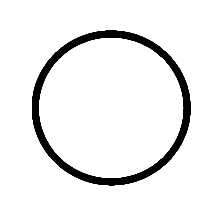}\includegraphics[scale=0.12]{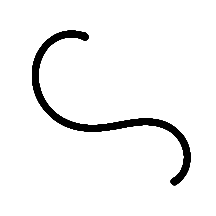}\includegraphics[scale=0.12]{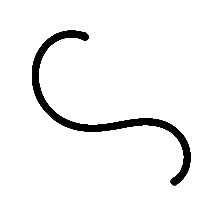} 
  \caption{Random Initialization}
  \label{fig_tSne_Random_Intialization_circle}
  \end{subfigure}%
  \begin{subfigure}{.24\textwidth}
  \centering
  \includegraphics[scale=0.12]{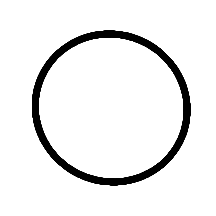}\includegraphics[scale=0.12]{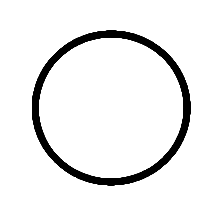}\includegraphics[scale=0.12]{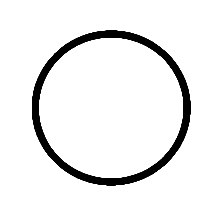}
  \caption{PCA Initialization}
  \label{fig_tSne_Informed_Intialization_circle}
  \end{subfigure}%

 \caption{t-SNE plots comparison for different kernel matrices with Random and PCA initialization for \textbf{Circle toy} dataset using Gaussian, Isolation, and Laplacian kernels, respectively.}
  \label{fig_tSne_Intialization_circle}
\end{figure}

This section reports the results of t-SNE for different kernels and initialization methods.
The deformed circular data with isolation and Laplacian kernel was created using the t-SNE method with the random initialization, also shown in Figure~\ref{fig_tSne_Random_Intialization_circle}. In this instance, it is clear that the overall structure was not maintained; in fact, gradient descent in t-SNE merely brings near neighbors closer and is hardly affected by the overall arrangement of points. While using PCA (frequently suggested for initialization in t-SNE), it restores the original circle and significantly enhances the t-SNE result (see Figure~\ref{fig_tSne_Informed_Intialization_circle}). While recovering the original circle for all $3$ kernels, informed initialization (PCA) simultaneously significantly improves the outcomes. Thus, it is clear that the closed one-dimensional manifold—the circle—can only be faithfully represented with informative initialization.


\begin{figure}[h!]
  \centering
  \begin{subfigure}{.16\textwidth}
  \centering
  \includegraphics[scale=0.5]{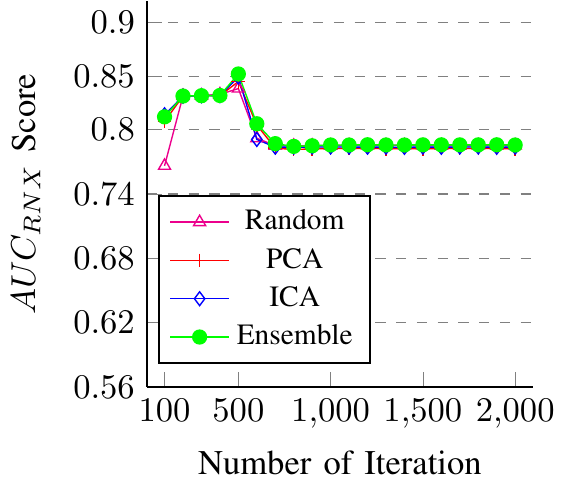}
  \caption{Gaussian Kernel}
  \end{subfigure}%
  \begin{subfigure}{.16\textwidth}
  \centering
  \includegraphics[scale=0.5]{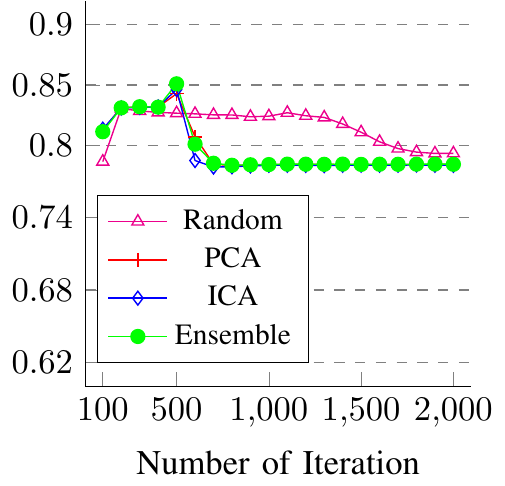}
  \caption{Isolation Kernel}
  \end{subfigure}%
  \begin{subfigure}{.16\textwidth}
  \centering
  \includegraphics[scale=0.5]{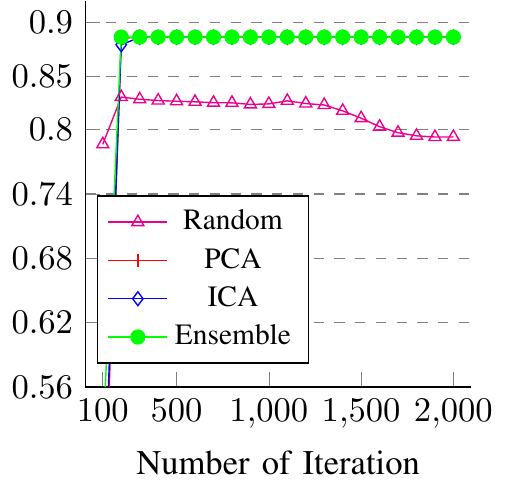}
  \caption{Laplacian Kernel}
  \end{subfigure}%
  \caption{AUC score of t-SNE for different Kernel methods applied with increasing number of iteration for \textbf{Circle data}. The figure is best seen in color.
  } 
  \label{fig_Auc_circle_data}
\end{figure}

In Figure~\ref{fig_Auc_circle_data}, using various kernels and initialization strategies, we present the $AUC_{RNX}$ charts for gradient descent iterations ranging from $100$ to $2000$. With ensemble initialization, the Laplacian Kernel outperforms the competition. Based on these results, we suggest a recommendation and avoiding combination in Table~\ref{tbl_recommended_result_summary}.

\begin{figure}[h!]
  \centering
  \begin{subfigure}{.24\textwidth}
  \centering
  \includegraphics[scale=0.055]{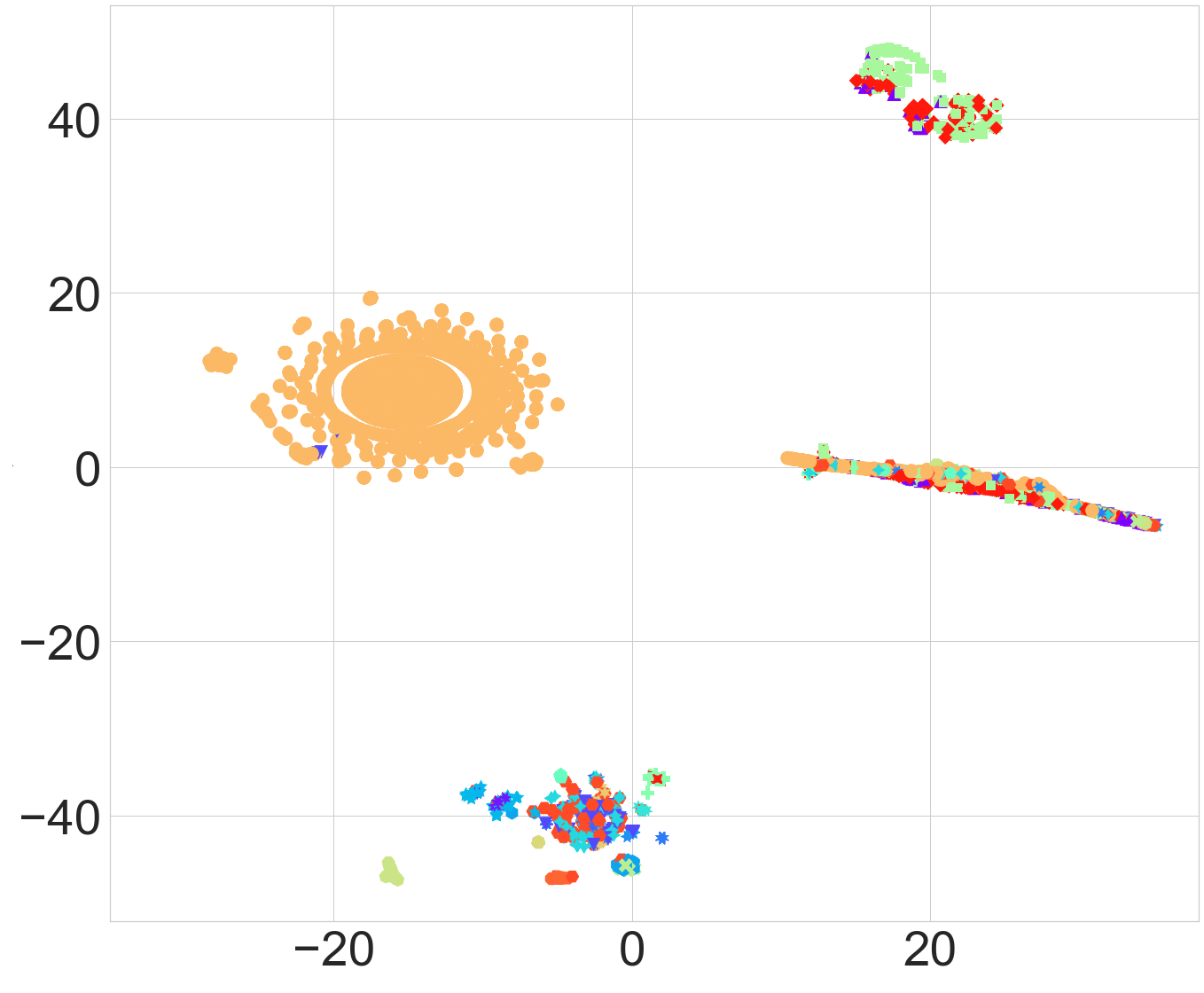}
  \caption{Gaussian}
  \end{subfigure}%
  \begin{subfigure}{.24\textwidth}
  \centering
 \includegraphics[scale=0.055]{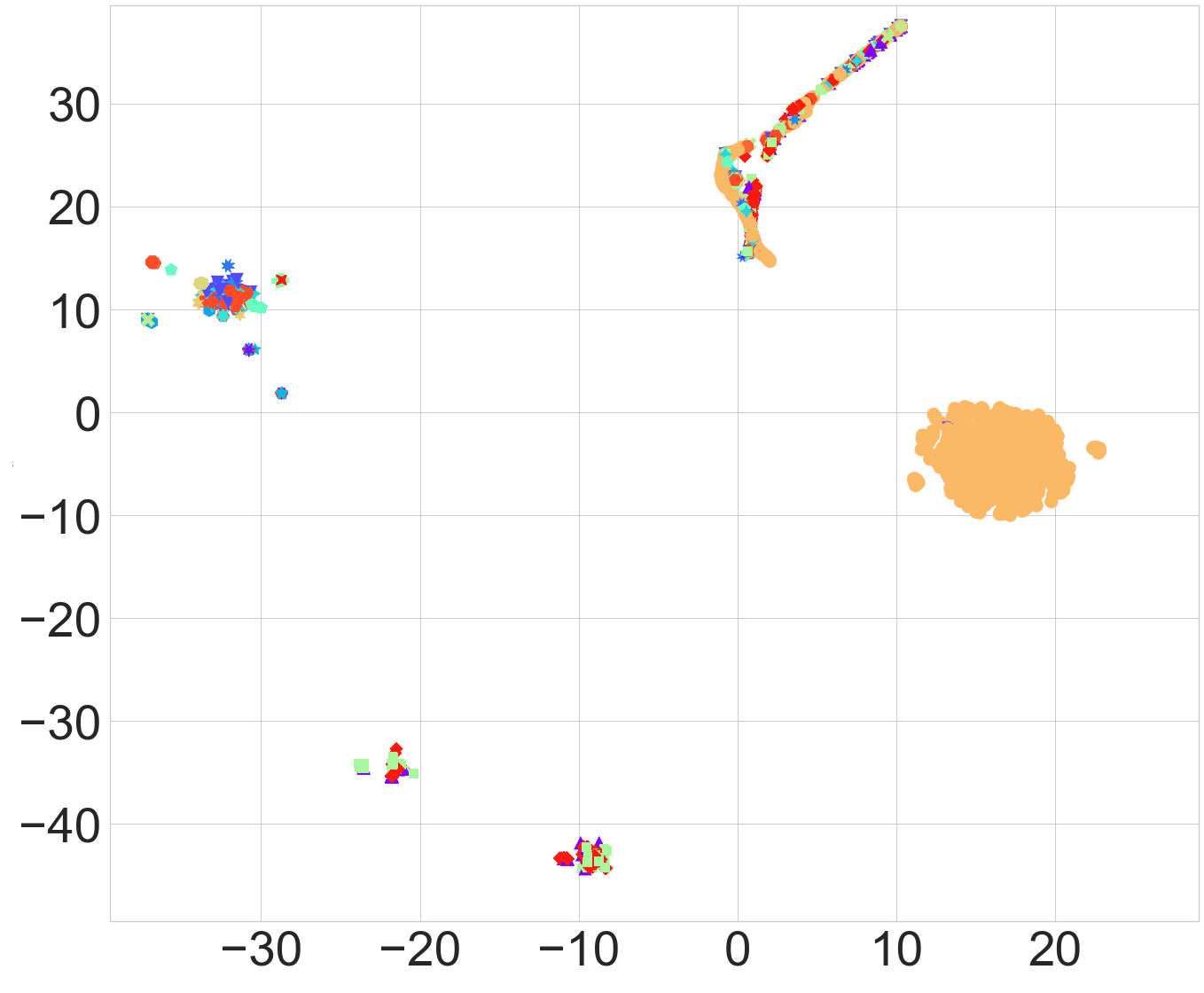}
  \caption{Isolation}
  \end{subfigure}%
  \\
  \begin{subfigure}{.24\textwidth}
  \centering
   \includegraphics[scale=0.055]{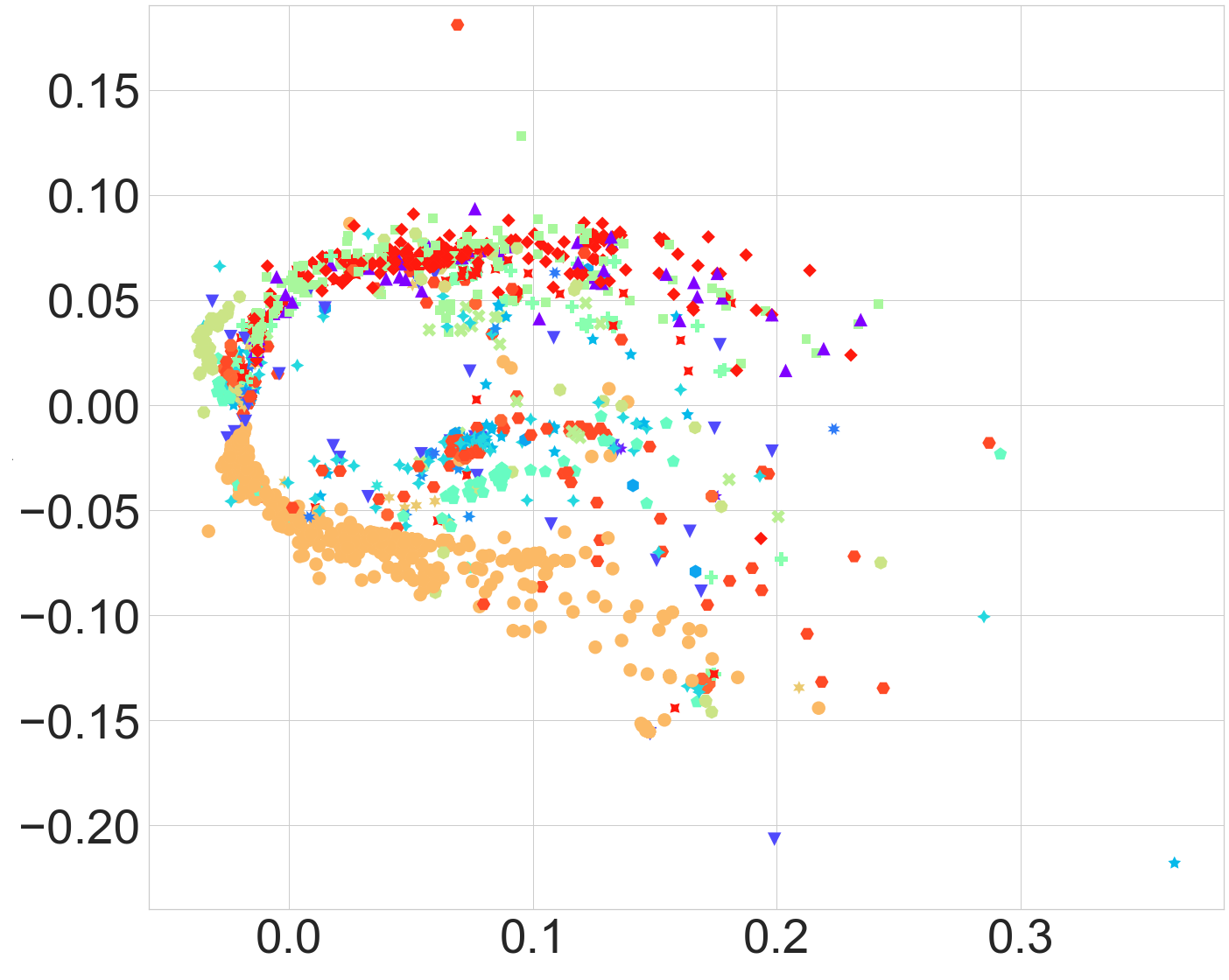} 
  \caption{Laplacian}
  \end{subfigure}%
  \begin{subfigure}{.24\textwidth}
  \centering
   \includegraphics[scale=0.055]{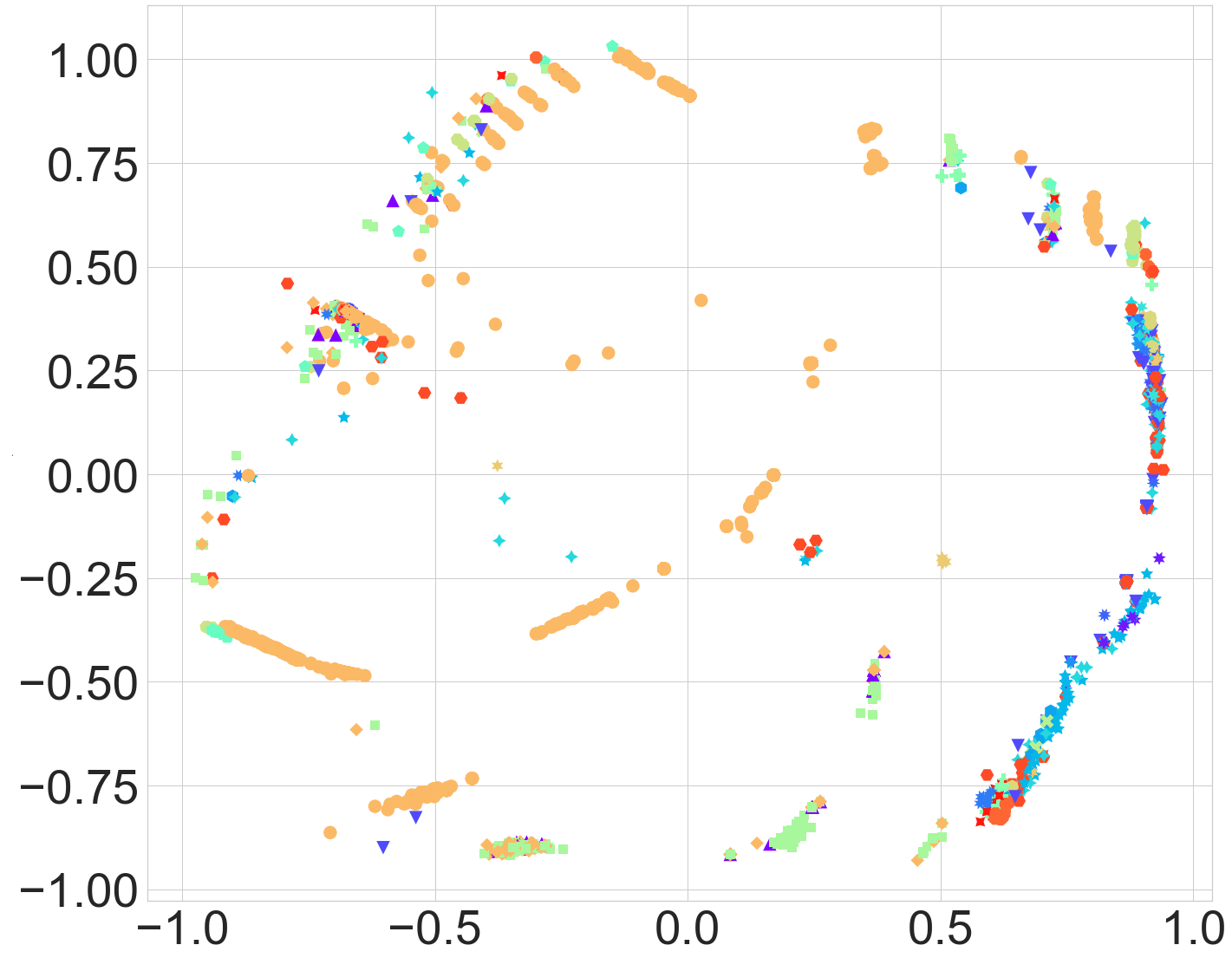} 
  \caption{Approximate}
  \end{subfigure}%
  \\
  \begin{subfigure}{0.5\textwidth}
  \centering
  \includegraphics[scale=0.15]{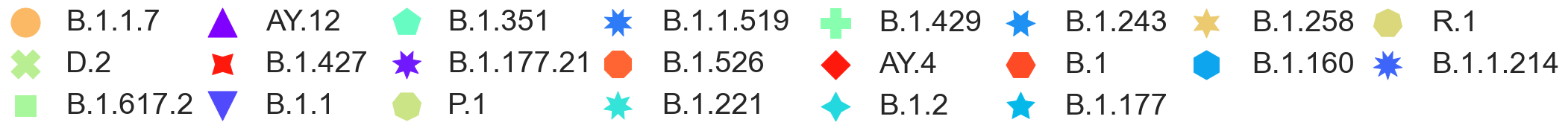}
  \end{subfigure}
   \caption{t-SNE plots for different kernel matrices with random initialization for \textbf{Spike Sequence} dataset.}
  \label{fig_tSne_Different_Intialization_spike}
\end{figure}

In Figure~\ref{fig_tSne_Different_Intialization_spike}, we report the t-SNE plots (for Spike sequence data) using the different kernels (for Spike2Vec-based embedding) and random initialization method. We can observe that the Alpha (B.1.1.7) variation displays unambiguous grouping in most cases.
For Gaussian and Isolation kernels, only the alpha variant is clearly separated. The other classes overlap with each other. However, for the Laplacian kernel, we can see smaller groups for other variants, such as Delta (AY.4) and Epsilon (B.1.429).

\begin{figure}[h!]
  \centering
  \begin{subfigure}{.24\textwidth}
  \centering
  \includegraphics[scale=0.48]{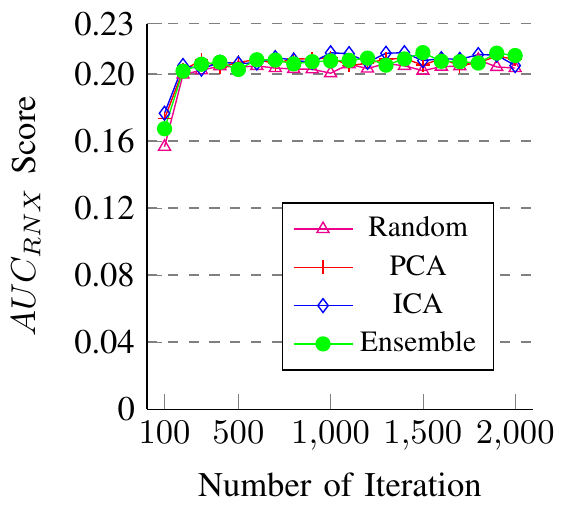}
  \caption{Gaussian Kernel}
  \end{subfigure}%
  \begin{subfigure}{.24\textwidth}
  \centering
  \includegraphics[scale=0.48]{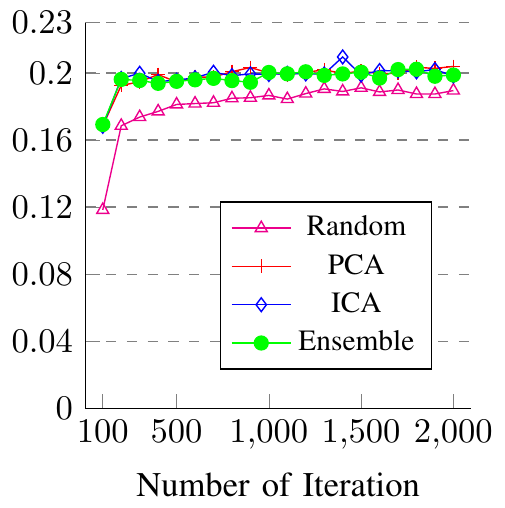}
  \caption{Isolation Kernel}
  \end{subfigure}%
  \\
  \begin{subfigure}{.24\textwidth}
  \centering
  \includegraphics[scale=0.48]{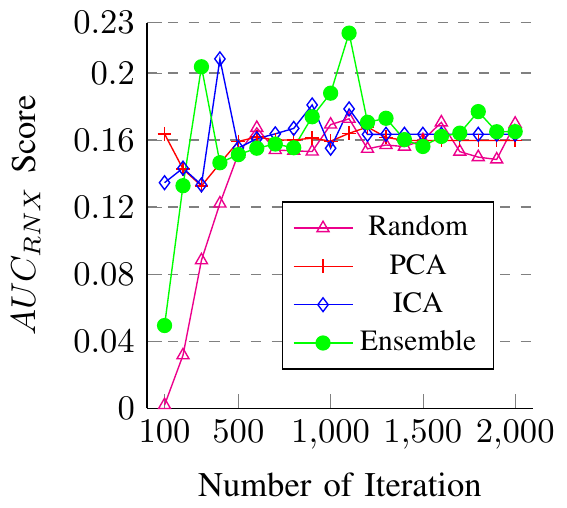}
  \caption{Laplacian Kernel}
  \end{subfigure}%
  \begin{subfigure}{.24\textwidth}
  \centering
  \includegraphics[scale=0.48]{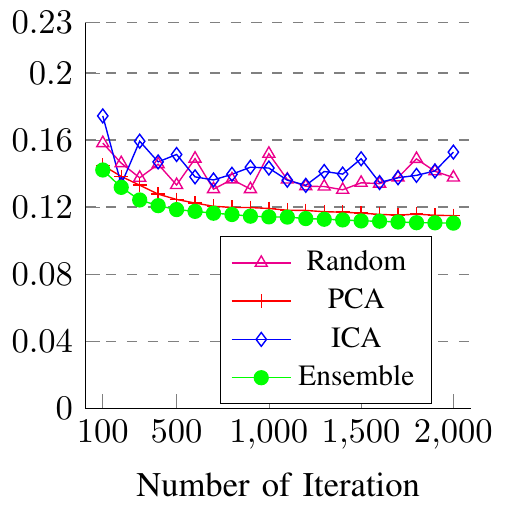}
  \caption{Approximate Kernel 
  }
  \end{subfigure}%
  \caption{AUC score of t-SNE for different Kernel methods and initialization with the increasing number of iterations for \textbf{Spike sequence data}. The figure is best seen in color.} 
  \label{fig_data_kernel_runtime}
\end{figure}

Figure~\ref{fig_data_kernel_runtime} shows the results of $AUC_{RNX}$ for different initialization approaches and kernel computation techniques for Spike data. For the Gaussian kernel, all initialization methods perform comparably to each other as the number of iterations increases. Similarly, for the Isolation kernel, we can observe that the random initialization-based approach is worse than other initialization methods. However, the other three (PCA, ICA, and Ensemble) methods seem to perform similarly. At around $1500^{th}$ iteration, the ICA-based initialization gets a spike in $AUC_{RNX}$ compared to the other methods. However, it is not significant as compared to an ensemble. For the Laplacian kernel, the ensemble initialization approach seems to perform better than the other methods, while random initialization performs worst at the start. Moreover, the maximum $AUC_{RNX}$ reported for the Laplacian kernel is the largest among all three kernels, which shows that the Gaussian kernel is the typical kernel used for t-SNE computation, may not be a good choice. Based on this result, We can recommend Laplacian with Ensemble and say Random performs poorly with Isolation and Laplacian Kernel, as mentioned in Table~\ref{tbl_recommended_result_summary}.

\begin{figure}[h!]
  \centering
  \begin{subfigure}{.24\textwidth}
  \centering
  \includegraphics[scale=0.055]{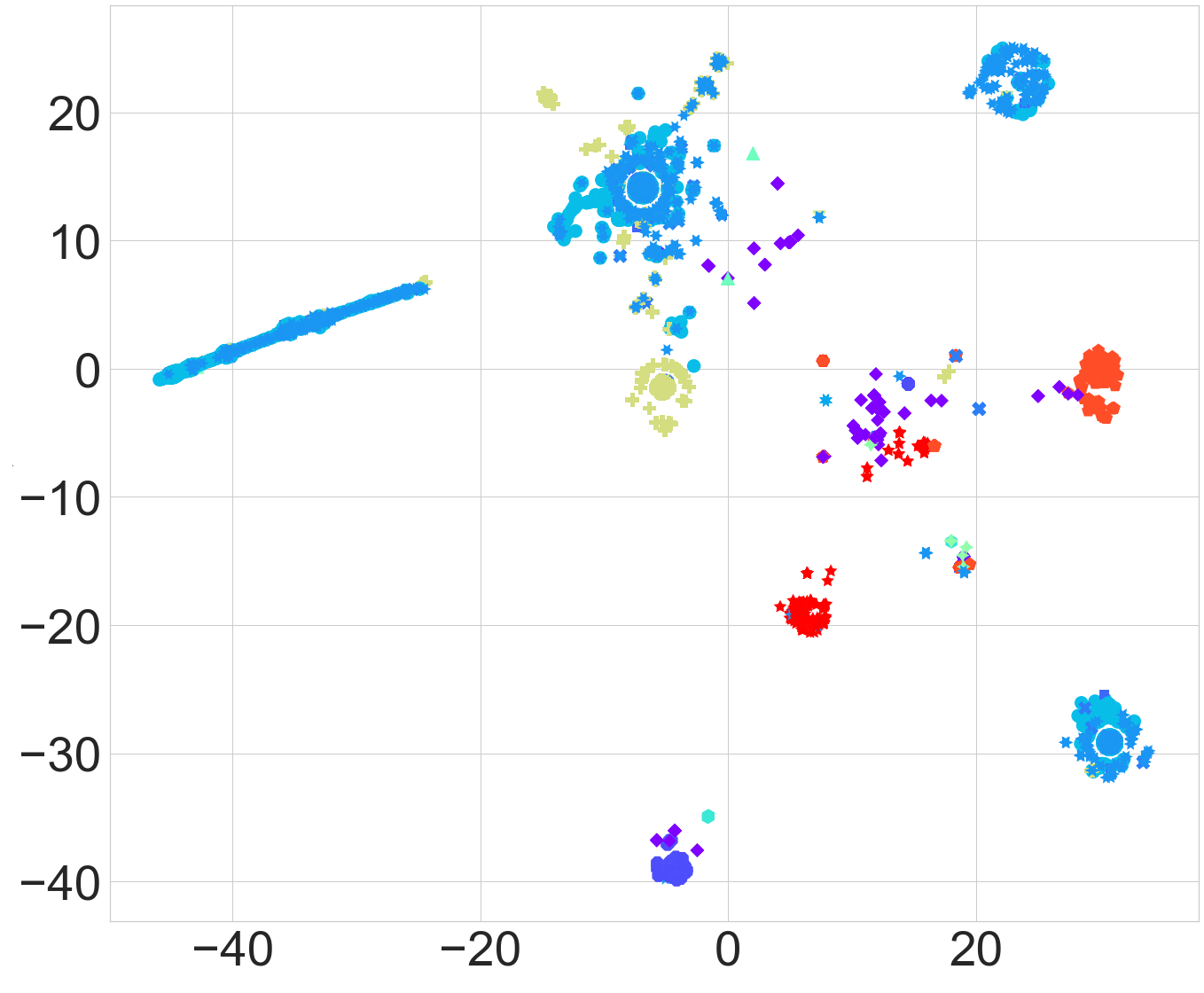}
  \caption{Gaussian}
  \end{subfigure}%
  \begin{subfigure}{.24\textwidth}
  \centering
 \includegraphics[scale=0.055]{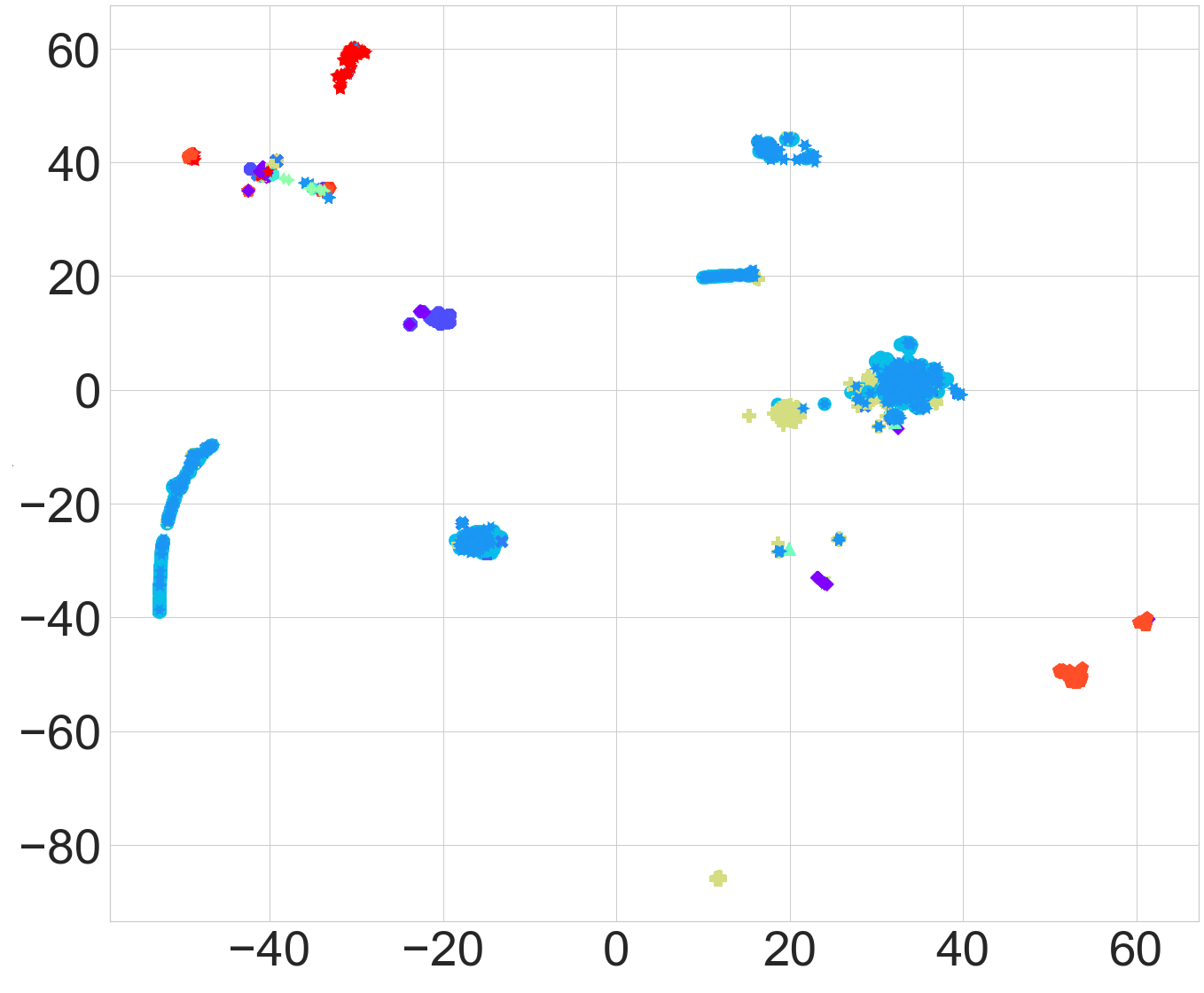}
  \caption{Isolation}
  \end{subfigure}%
  \\
  \begin{subfigure}{.24\textwidth}
  \centering
   \includegraphics[scale=0.055]{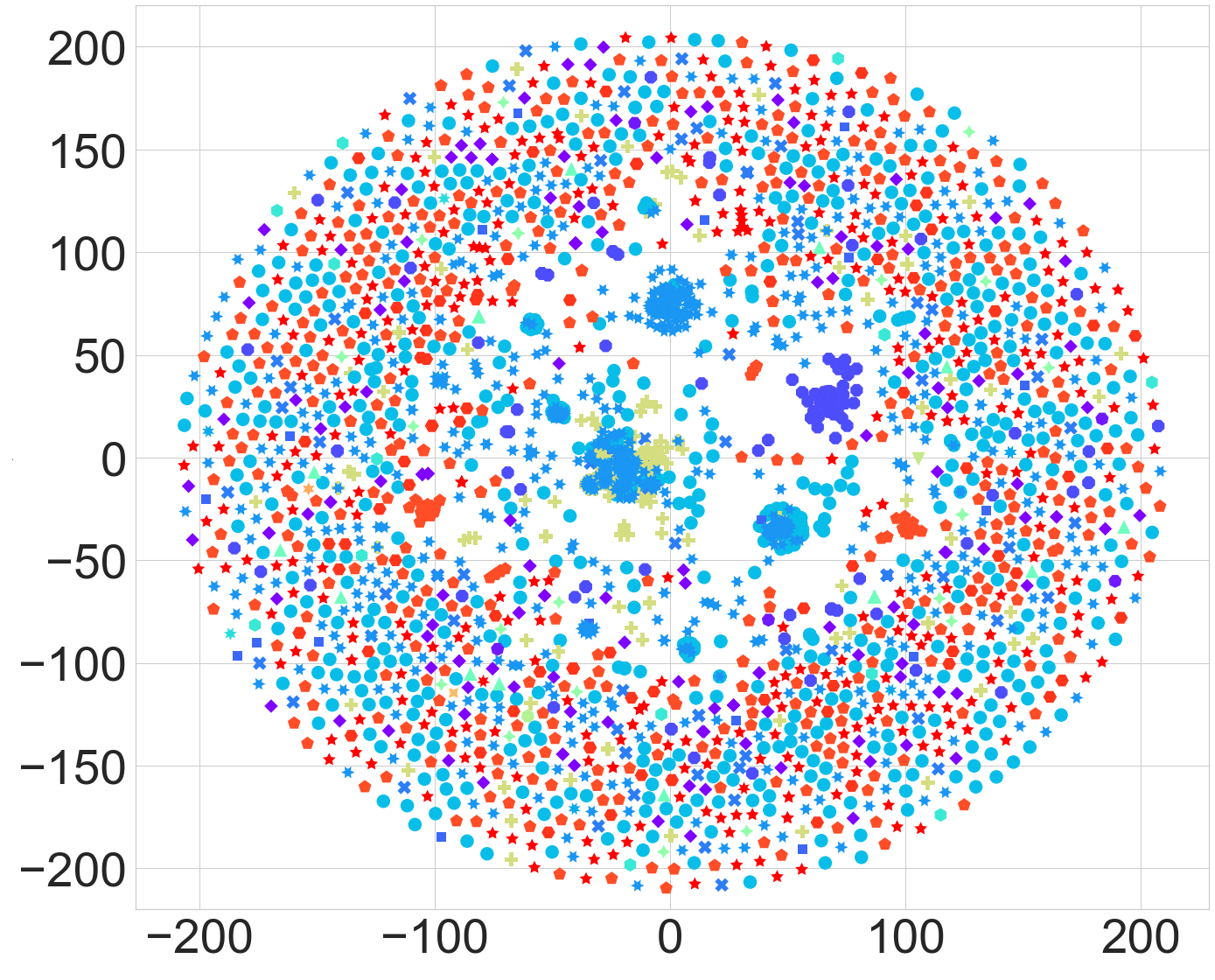} 
  \caption{Laplacian}
  \end{subfigure}%
  \begin{subfigure}{.24\textwidth}
  \centering
   \includegraphics[scale=0.055]{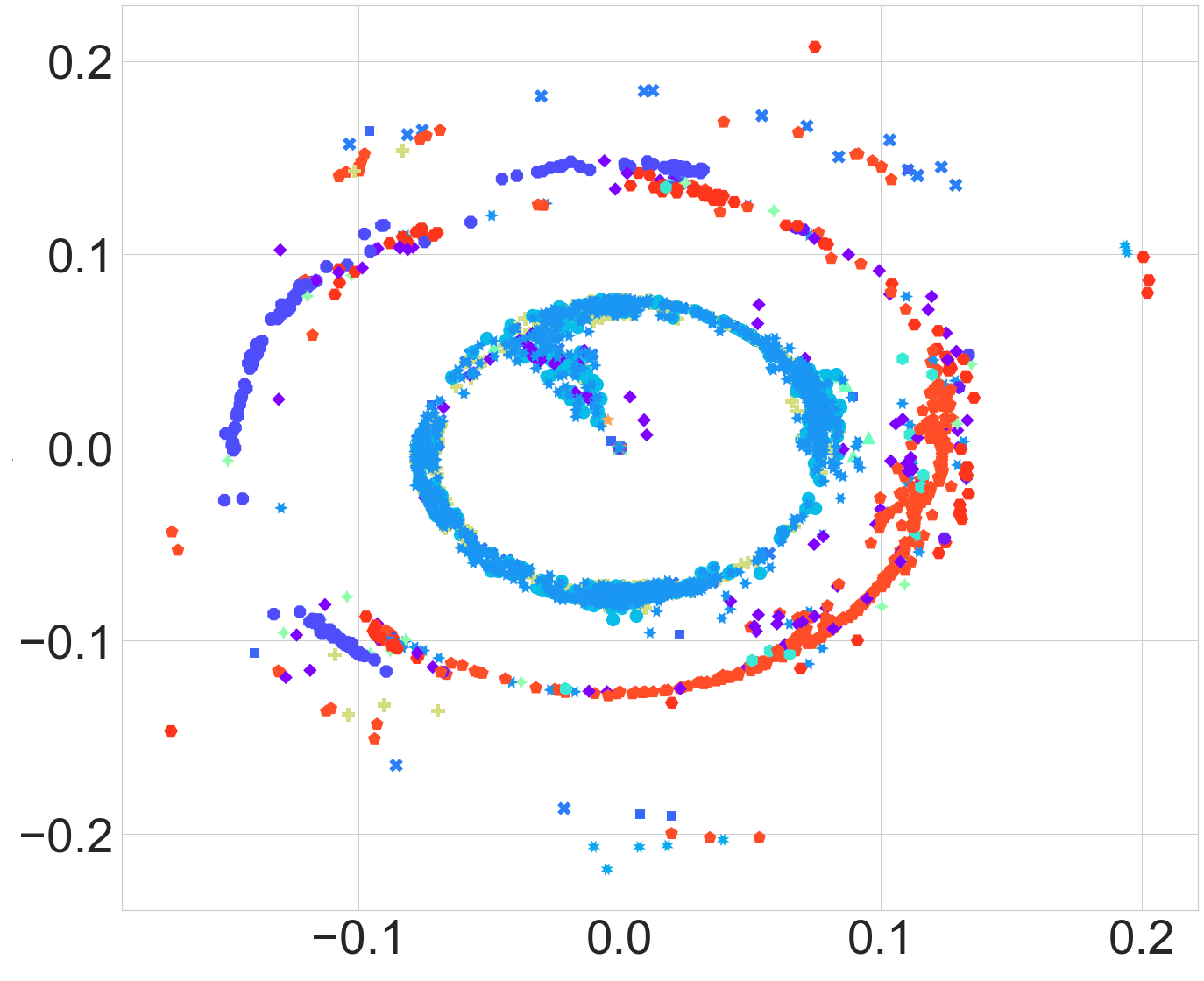} 
  \caption{Approximate}
  \end{subfigure}%
  \\
  \begin{subfigure}{0.5\textwidth}
  \centering
  \includegraphics[scale=0.16]{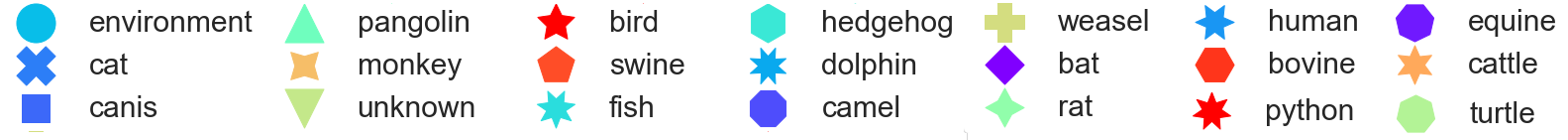}
  \end{subfigure}
   \caption{t-SNE plots for different kernel matrices with random initialization for \textbf{Coronavirus Host} dataset.}
  \label{fig_tSne_Different_Intialization_host}
\end{figure}

In Figure~\ref{fig_tSne_Different_Intialization_host}, we report the t-SNE plots using the different kernels and random initialization method for Host data. In most cases, the Environment and Human displays unambiguous grouping. 
For Laplacian, classes overlap with each other. However, for the Approximate kernel, we can see smaller groups for other variants, such as Bird and Swine.

\begin{figure}[h!]
  \centering
  \begin{subfigure}{.24\textwidth}
  \centering
  \includegraphics[scale=0.48]{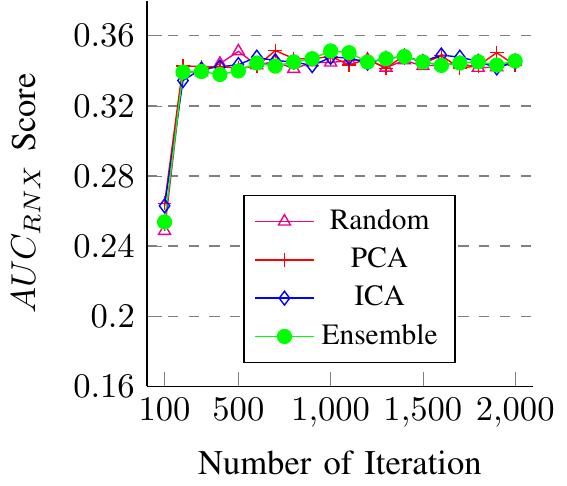}
  \caption{Gaussian Kernel}
  \end{subfigure}%
  \begin{subfigure}{.24\textwidth}
  \centering
  \includegraphics[scale=0.48]{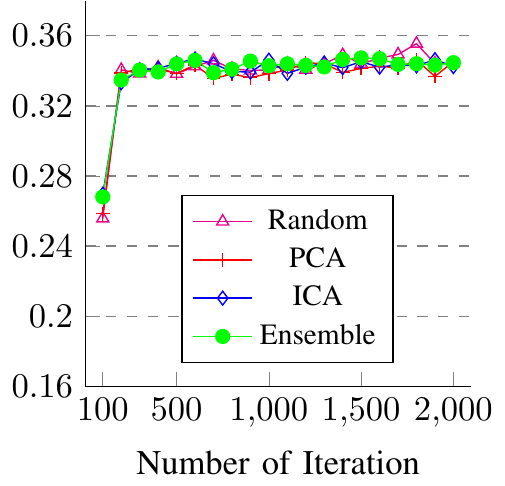}
  \caption{Isolation Kernel}
  \end{subfigure}%
  \\
  \begin{subfigure}{.24\textwidth}
  \centering
  \includegraphics[scale=0.48]{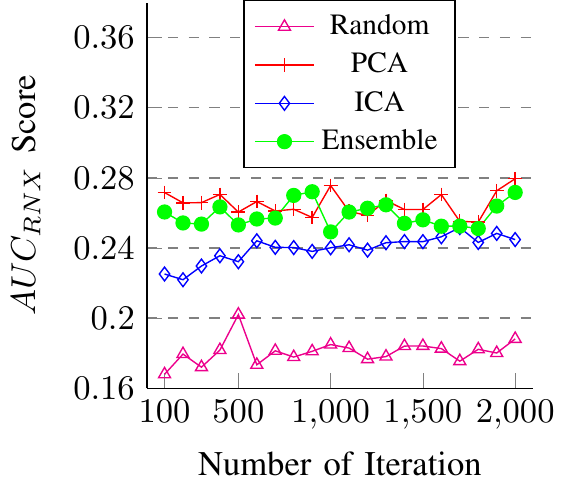}
  \caption{Laplacian Kernel}
  \end{subfigure}%
  \begin{subfigure}{.24\textwidth}
  \centering
  \includegraphics[scale=0.48]{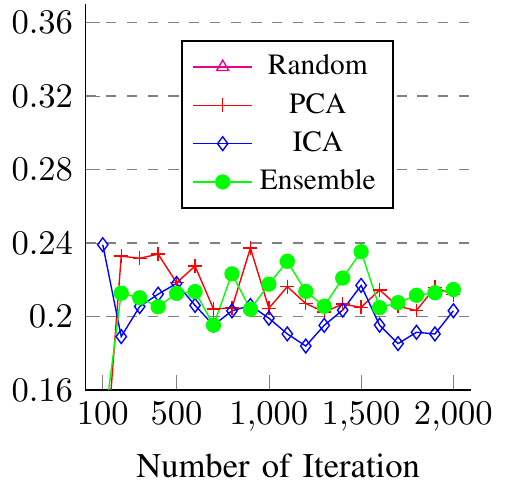}
  \caption{Approximate Kernel}
  \end{subfigure}%
  \caption{AUC score of t-SNE for different Kernel with an increasing number of iterations for \textbf{Coronavirus Host} data comprised of 5558 sequences. The figure is best seen in color.} 
  \label{fig_Auc_host_data}
\end{figure}

Figure~\ref{fig_Auc_host_data} shows the results of $AUC_{RNX}$ for different initialization approaches and kernel computation techniques for Host data. For the Gaussian kernel and Isolation Kernel, all initialization methods perform comparably to each other as the number of iterations increases. In comparison, Laplacian and Approximate are worse and can not be compared. However, among them, random initialization significantly performs worst than other initialization in Laplacian. Ensemble methods seem comparable in general (especially with Approximate kernel). Overall Isolation kernel, with the ensemble initialization approach, seems to perform better than the other methods. In contrast, random initialization with Laplacian and ICA initialization with Approximate Kernel perform worst. The summary is shown in Table~\ref{tbl_recommended_result_summary}.


\begin{figure}[h!]
  \centering
  \begin{subfigure}{.24\textwidth}
  \centering
  \includegraphics[scale=0.055]{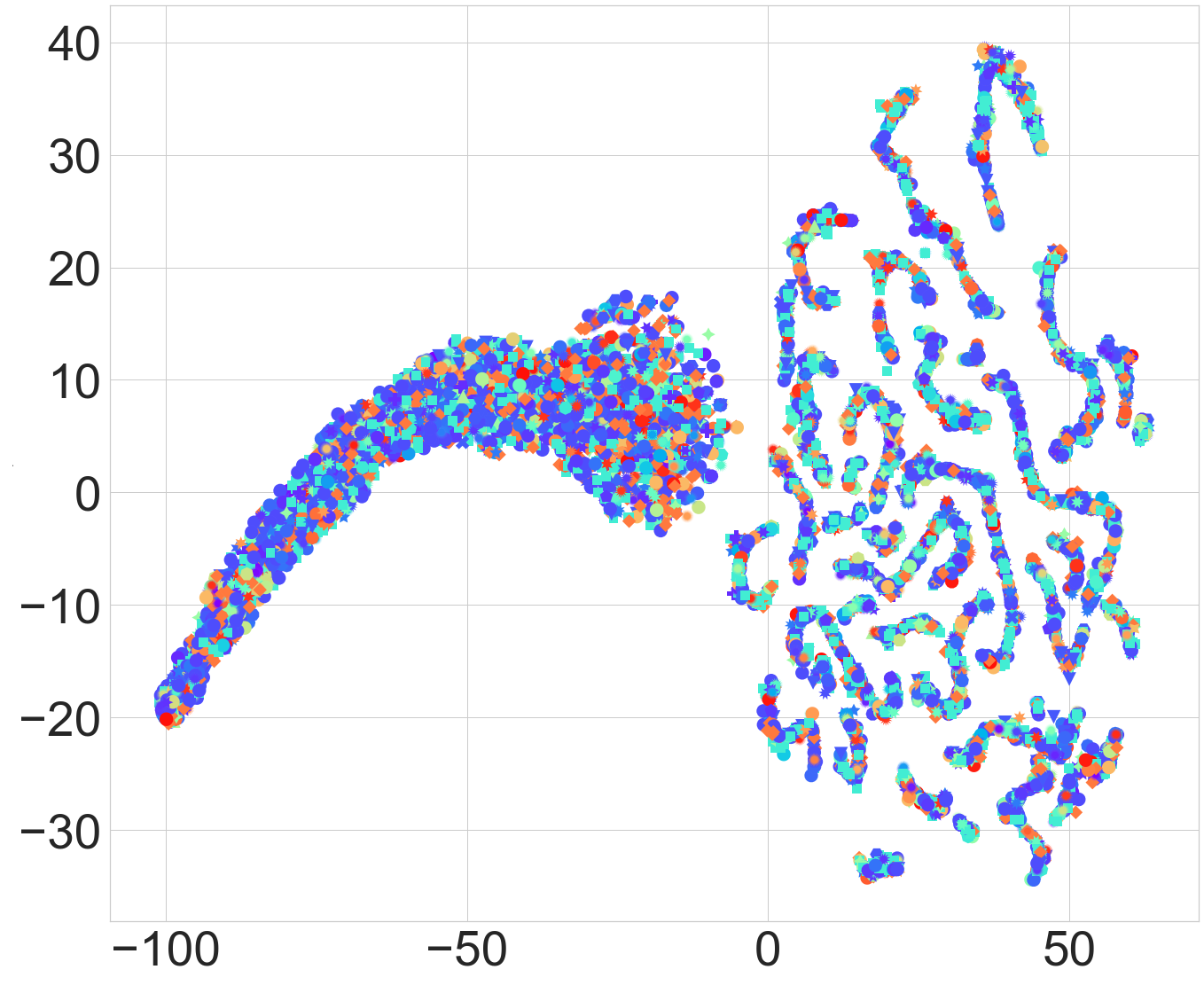}
  \caption{Gaussian}
  \end{subfigure}%
  \begin{subfigure}{.24\textwidth}
  \centering
 \includegraphics[scale=0.055]{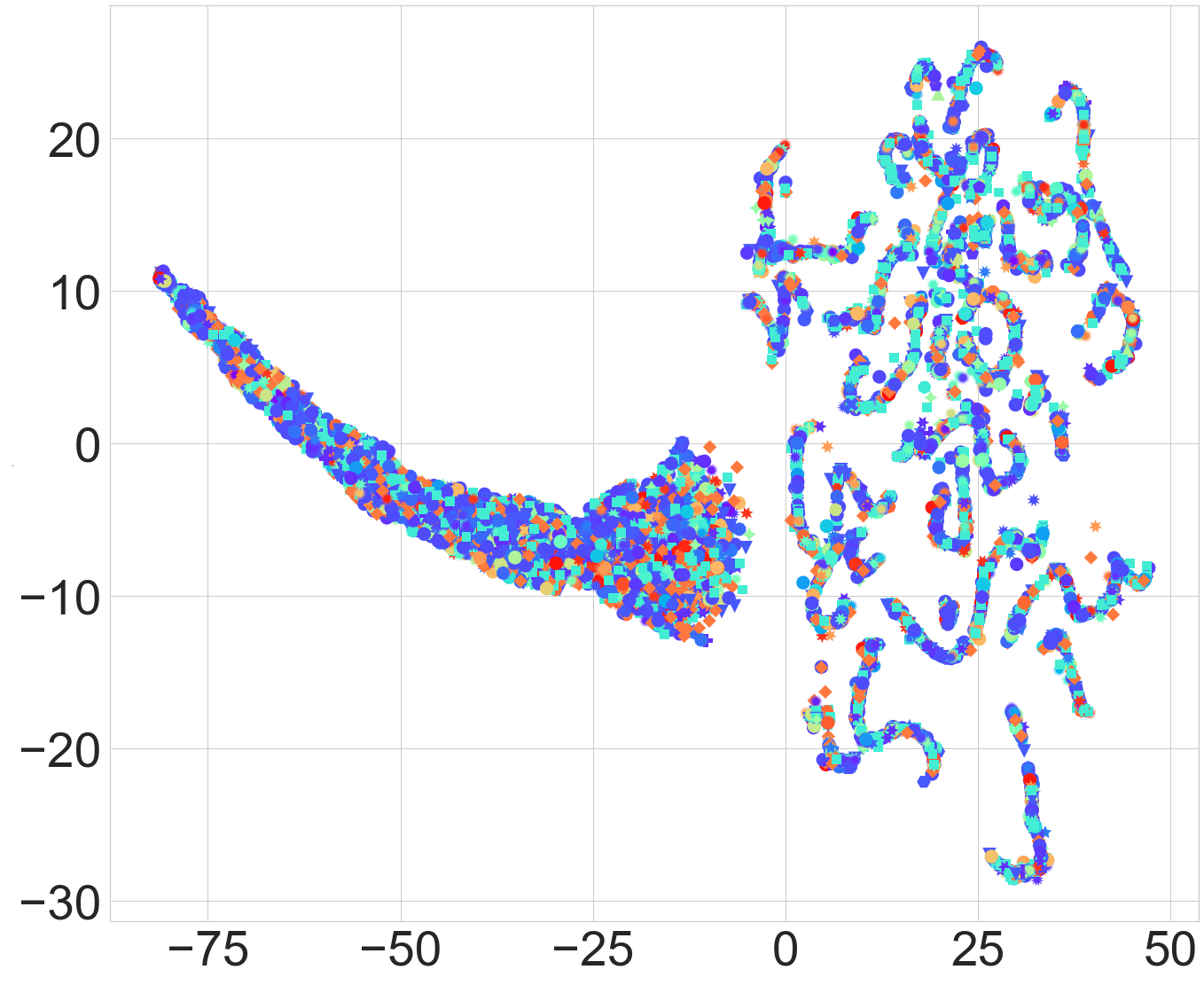}
  \caption{Isolation}
  \end{subfigure}%
  \\
  \begin{subfigure}{.24\textwidth}
  \centering
   \includegraphics[scale=0.055]{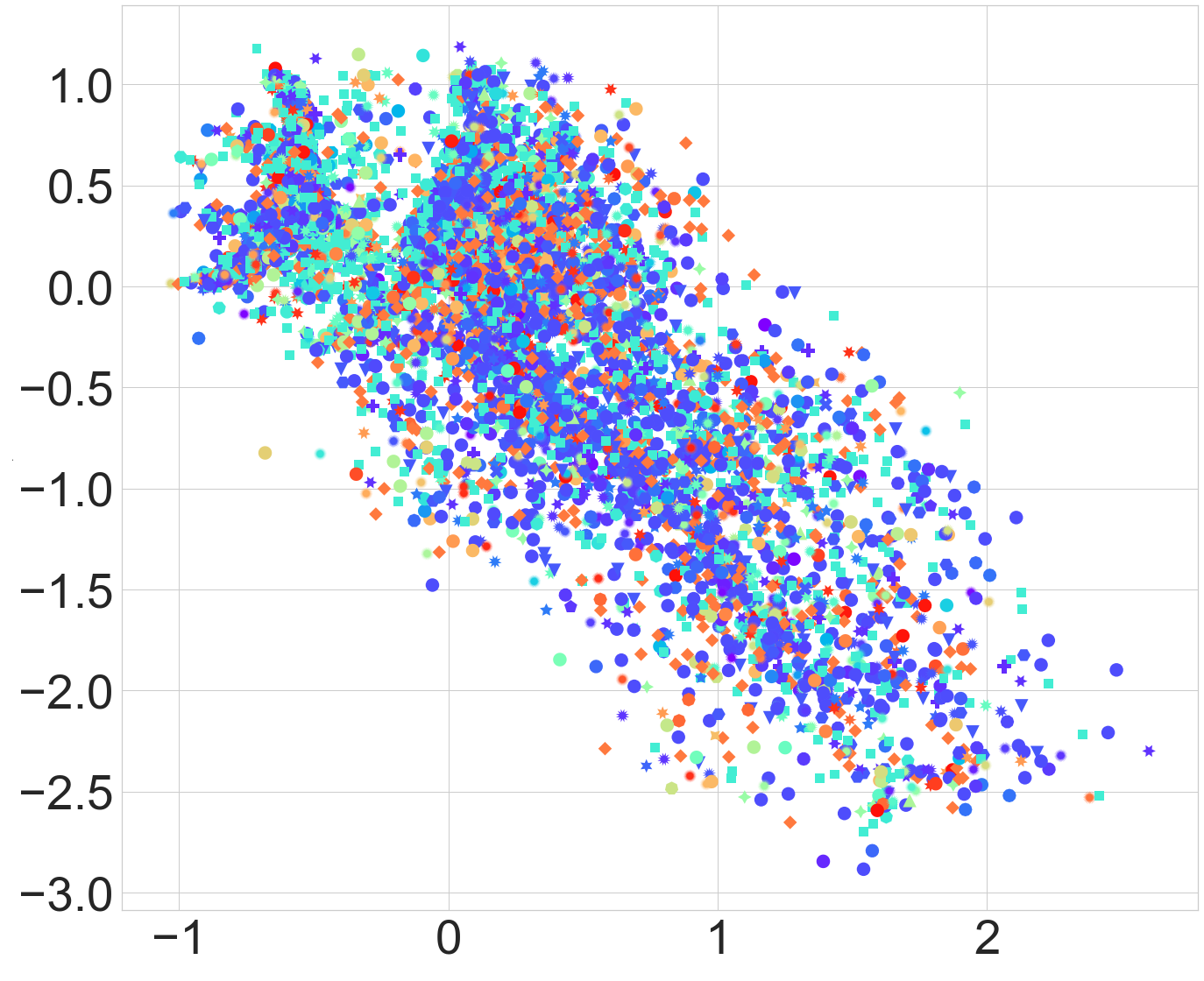} 
  \caption{Laplacian}
  \end{subfigure}%
  \begin{subfigure}{.24\textwidth}
  \centering
   \includegraphics[scale=0.055]{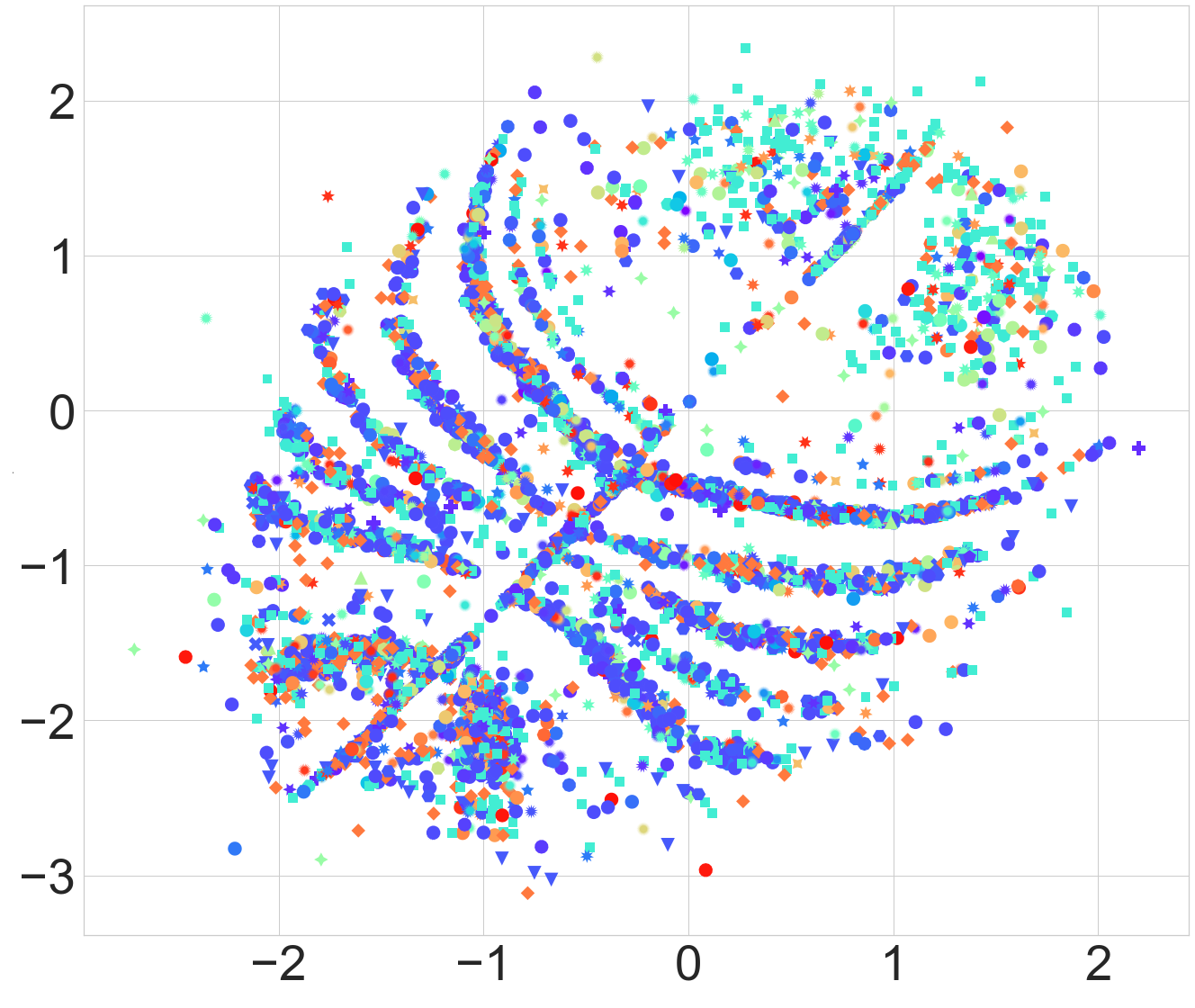} 
  \caption{Approximate}
  \end{subfigure}%
   \caption{t-SNE plots for different kernels with random initialization for \textbf{Short Read} dataset. The legends for $496$ classes are not shown  to save space. The figure is best seen in color.   
   }
  \label{fig_tSne_Different_Intialization_shortRead}
\end{figure}

In Figure~\ref{fig_tSne_Different_Intialization_shortRead}, we report the t-SNE plots using the different kernels (for Spike2Vec-based embedding) and random initialization method for Short Read data. We can see Gaussian and Isolation are somewhat similar, but Laplacian and Approximate are overlapping, and hard to find groups in them. A high number of classes($496$) is one of the reasons. 

\begin{figure}[h!]
  \centering
  \begin{subfigure}{.24\textwidth}
  \centering
  \includegraphics[scale=0.48]{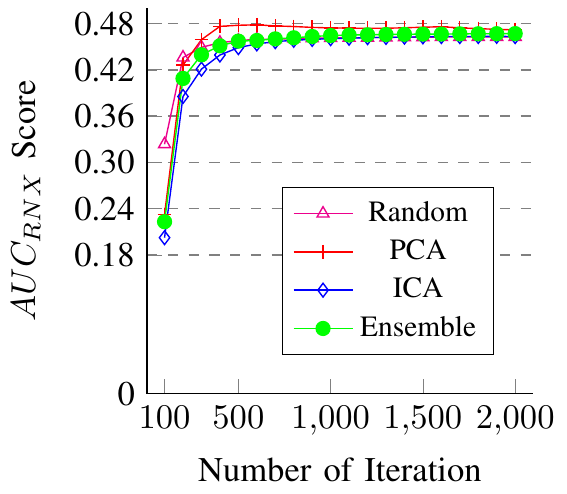}
  \caption{Gaussian Kernel}
  \end{subfigure}%
  \begin{subfigure}{.24\textwidth}
  \centering
  \includegraphics[scale=0.48]{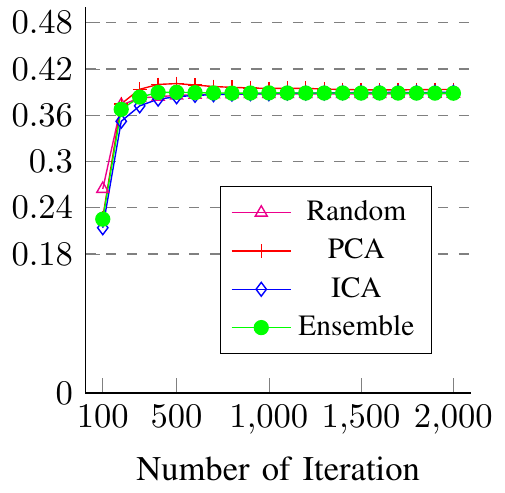}
  \caption{Isolation Kernel}
  \end{subfigure}%
  \\
  \begin{subfigure}{.24\textwidth}
  \centering
  \includegraphics[scale=0.48]{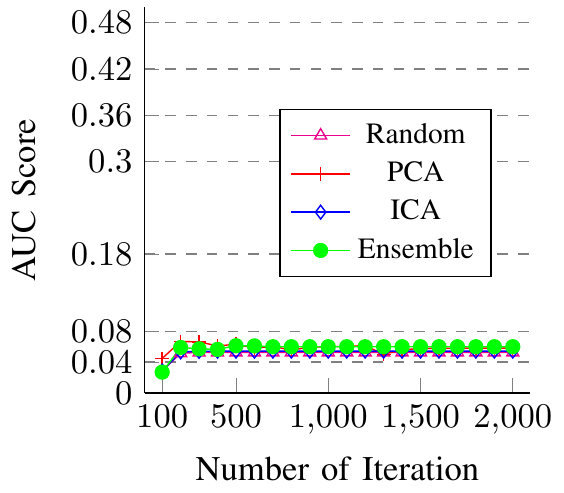}
  \caption{Laplacian Kernel}
  \end{subfigure}%
  \begin{subfigure}{.24\textwidth}
  \centering
  \includegraphics[scale=0.48]{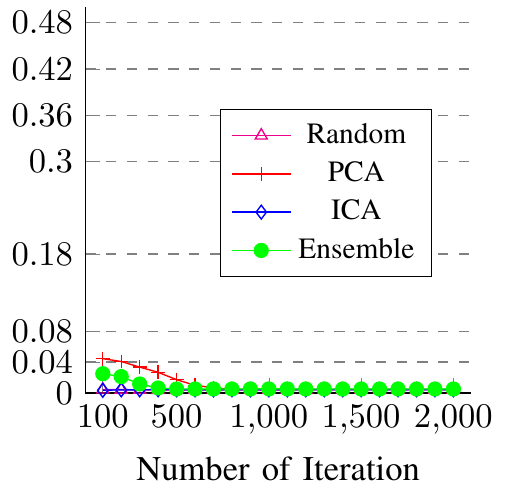}
  \caption{Approximate Kernel 
  }
  \end{subfigure}%
  \caption{AUC score of t-SNE for different Kernel methods and initialization with the increasing number of iterations for \textbf{Short Read data}. The figure is best seen in color.} 
  \label{fig_Auc_shortRead_data}
\end{figure}

Figure~\ref{fig_Auc_shortRead_data} shows the results of $AUC_{RNX}$ for Short Read data. For the Gaussian kernel, all initialization methods perform comparably to each other as the number of iterations increases. Isolation is somewhat close but incomparable with Gaussian, whereas Laplacian and Approximate Kernels are significantly worse. However, among all initialization, Ensemble methods seem to perform better. Overall Gaussian kernel, with the ensemble initialization approach, appears to be a reasonable choice. Conversely, any other Kernel seems not a good choice, as mentioned in the recommendation summary in Table~\ref{tbl_recommended_result_summary}. 


\begin{table}[h!]
\centering
\resizebox{0.35\textwidth}{!}{
    \begin{tabular}{c@{\extracolsep{4pt}}ccccc}
      \toprule
      \multirow{2}{*}{\raisebox{-\heavyrulewidth}{Dataset}} & \multicolumn{2}{c}{Recommended} & \multicolumn{2}{c}{Not Recommended} \\
      \cmidrule{2-3}  \cmidrule{4-5}
      & Kernel & Initialization  & Kernel & Initialization  \\
      \midrule \midrule
      \multirow{2}{*}{Circle} & \multirow{2}{*}{Laplacian} & \multirow{2}{*}{Ensemble} & Isolation & ICA  \\
      & & & Laplacian & Random \\
      \cmidrule{2-5}
      \multirow{3}{*}{Spike} & \multirow{3}{*}{Laplacian} & \multirow{3}{*}{Ensemble} & Approximate &  PCA  \\
      & & & Isolation &  Random  \\
      & & & Laplacian &  Random  \\
      \cmidrule{2-5}
      \multirow{2}{*}{Host} & \multirow{2}{*}{Isolation} & \multirow{2}{*}{Ensemble} & Approximate & ICA  \\
      & & & Laplacian & Random  \\
      \cmidrule{2-5}
      \multirow{1}{*}{Short Read} & Gaussian & Ensemble & All others & None  \\
      \bottomrule
    \end{tabular}
    }
    \caption{Recommendation based on summary of performance ($AUC_{RNX}$) on different dataset.}
    \label{tbl_recommended_result_summary}
\end{table}

\section{Conclusion}\label{sec_conclusion}
We propose using an ensemble initialization procedure to improve the performance of t-SNE for biological sequences. We show that kernel selection can also play a crucial role along with ensemble initialization to improve the performance of t-SNE. In the future, we will explore more kernels and initialization methods along with other biological data to study the behavior of t-SNE.

\bibliographystyle{IEEEtran}
\bibliography{references}

\end{document}